\documentclass[nohyperref]{article}
\pdfoutput=1
\usepackage{microtype}
\usepackage{graphicx}
\usepackage{booktabs} 
\usepackage{caption, subcaption}

\usepackage{hyperref}

\usepackage[accepted]{icml2022}

\usepackage{amsmath}
\usepackage{amssymb}
\usepackage{mathtools}
\usepackage{amsthm}

\usepackage[capitalize,noabbrev]{cleveref}

\theoremstyle{plain}

\theoremstyle{definition}

\theoremstyle{remark}

\usepackage[textsize=tiny]{todonotes}

\icmltitlerunning{Monitoring shortcut learning using mutual information}

\begin{document}

\twocolumn[
\icmltitle{Monitoring Shortcut Learning using Mutual Information}





\begin{icmlauthorlist}
\icmlauthor{Mohammed~Adnan}{aaa,bbb}
\icmlauthor{Yani~Ioannou}{ccc}
\icmlauthor{Chuan-Yung~Tsai}{bbb}
\icmlauthor{Angus~Galloway}{aaa,bbb}\\
\icmlauthor{H.R.~Tizhoosh}{ddd,eee}
\icmlauthor{Graham~W.~Taylor}{aaa,bbb}

\end{icmlauthorlist}

\icmlaffiliation{aaa}{University of Guelph}
\icmlaffiliation{bbb}{Vector Institute, Canada}
\icmlaffiliation{ccc}{University of Calgary}
\icmlaffiliation{ddd}{Mayo Clinic, Rochester, MN, USA}
\icmlaffiliation{eee}{Kimia Lab, University of Waterloo, Canada}

\icmlcorrespondingauthor{\mbox{Mohammed~Adnan}}{madnan01@uoguelph.ca}
\icmlcorrespondingauthor{\mbox{Graham~Taylor}}{gwtaylor@uoguelph.ca}

\icmlkeywords{shortcuts, spurious correlation, information bottleneck}

\vskip 0.3in
]



\printAffiliationsAndNotice{}  

\begin{abstract}

The failure of deep neural networks to generalize to out-of-distribution data is a well-known problem and raises concerns about the deployment of trained networks in safety-critical domains such as healthcare, finance and autonomous vehicles. We study a particular kind of distribution shift --- \emph{shortcuts} or \emph{spurious correlations} in the training data. Shortcut learning is often only exposed when models are evaluated on real-world data that does not contain the same spurious correlations, posing a serious dilemma for AI practitioners to properly assess the effectiveness of a trained model for real-world applications. In this work, we propose to use the mutual information (MI) between the learned representation and the input as a metric to find where in training the network latches onto shortcuts. 
Experiments demonstrate that MI can be used as a domain-agnostic metric for monitoring shortcut learning.
  

\end{abstract}

\section{Introduction}
\label{intro}
Our understanding of `how' and `what' neural networks learn is limited, which raises concern about the deployment of neural networks in safety-critical domains.
Despite achieving state-of-the-art performance on benchmark datasets, neural networks may fail to generalize in real-world settings or for out-of-distribution data~\cite{koh2021wilds}. For example, models trained for cancer detection may not generalize on data from a new hospital~\cite{Castro2020CausalityMI, perone_domain, albadawy2018deep} and self-driving cars may not generalize to new lighting conditions or object poses~\cite{alcorn,dai_van_gool}.
One reason why models may fail in real-world settings could be attributed to learning \emph{shortcuts}~\citep{Geirhos2020-dh} from the training data. A \emph{shortcut} is a type of distribution shift where spurious correlations exist only in the training data, resulting in the learning of non-intended or easy-to-learn discriminatory features which work well on the training and test dataset but not on out-of-distribution real-world datasets~\cite{wiles2022a, Geirhos2020-dh}. Shortcuts can arise due to dataset biases or the model using `trivial' or unintended features like high-frequency noise patterns or the image background for the classification task. For example, an Inception-V3 model trained to detect hip fractures was found to use scanner information for learning discriminatory features~\cite{Badgeley2019DeepLP}; deep learning systems trained to detect COVID-19 from chest radiographs can rely on confounding factors (shortcuts) rather than medical pathology~\cite{degrave2021ai}. While our understanding of shortcuts and how they arise is still developing, a helpful tool to practitioners deploying machine learning models in safety-critical domains with a high cost of failure would be to monitor shortcuts during the training phase. Although the phenomenon of shortcut learning is widely known, there is no effective method available to monitor shortcuts being learned. Interpretable machine-learning methods such as feature attribution, Grad-CAM~\cite{selvaraju2017grad}, and LIME~\cite{lime_rib} have been used to understand a model's dependency on spurious correlations. However, it has been shown that such post-hoc explanations are ineffective~\cite{adebayo2021post,  alqaraawi2020evaluating, chu2020visual}.\looseness=-1

In this work, we show that shortcut learning can be understood using the information bottleneck framework~\cite{tishby1999information, tishby2015deep}, by using the mutual information (MI) between the inputs and the learned representation to monitor shortcut learning. We use the neural tangent kernel (NTK)~\citep{Jacot2018-ha} to study the training evolution of shortcut learning. We design experiments using synthetic and complex real-world data to demonstrate the relationship between (MI) and shortcut learning, and show that MI can be used as a metric for domain-agnostic assessment of shortcuts. We find that compression as measured by MI is associated with the tendency to learn shortcuts.
    

\section{Background}
\label{background}

\paragraph{Shortcut learning:} 
\Citet{wiles2022a} defined shortcuts or spurious correlations as a type of \emph{distribution shift} such that two or more attributes are correlated at training time, but not for the test data, where they are independent. In a more general sense, shortcuts are \emph{easy-to-learn} decision rules that can be exploited in the absence of distribution shift (i.e.~on standard benchmarks) but fail to transfer to more challenging and diverse testing conditions, such as real-world datasets~\citep{Geirhos2020-dh}. 

\paragraph{Information bottleneck method:}
The information bottleneck (IB) can be viewed as a rate-distortion problem cast entirely in terms of mutual
information (MI)---denoted ``$I(V; Z)=I(Z; V)$''~\citep{tishby1999information}. A~\emph{distortion} function measures how well a relevant  variable $V$ is predicted from another variable $Z$, where $Z$ is usually a compressed 
representation of the input $X$. The~\emph{rate} refers to the complexity of $Z$, which is less than or equal to
$X$. IB is a general method for data compression but has been advanced as an explanatory tool predictive of
learning and generalization of neural networks (NNs). It has been suggested that NNs trained by SGD may learn compressed representations $Z$ of their input, making them insensitive to data idiosyncrasies, yet maintain sufficient~\emph{relevant} information for predicting 
the output $Y$ (e.g.~class labels)~\citep{tishby2015deep, shwartz2017opening}. 
This trade-off between compression and preserving task relevant information is optimized by the 
notion of ``minimal sufficient statistics''~\citep{cover1991elements}. 
The IB view suggests that NNs trained by cross-entropy loss may implicitly minimize the following Lagrangian:
\begin{equation}
    \min I(X; Z) - \beta I(Z; Y),
\end{equation}
enabling them to implement minimal sufficient statistics for different $\beta$-constraints on the error.\footnote{To what extent the relationship between IB and deep learning holds in general is 
the subject of ongoing debate~\citep{saxe2018information, jacobsen2018irevnet, goldfeld2019estimating}.}

\paragraph{Neural tangent kernel:}
Distribution-free estimation of MI for high-dimensional data is challenging and often intractable. One workaround to this
problem is using an infinite ensemble of infinite-width neural networks~\citep{pmlr-v118-shwartz-ziv20a} which
confer tractable bounds on MI.
The neural tangent kernel (NTK)~\citep{Jacot2018-ha} is a kernel that describes the evolution of infinite-width neural networks during their training by gradient descent, thus allowing the systematic study of neural networks using tools from kernel methods. Infinite-width neural networks behave as linear functions, and their training evolution can be fully described by the NTK. 
\citet{pmlr-v118-shwartz-ziv20a} used the fact that the output of an infinite ensemble of infinitely-wide neural networks initialized with Gaussian weights and biases and trained by MSE loss is a conditional Gaussian distribution. This allows tractable computation of (MI) between the representation $Z$ and the targets $Y$: $I(Z; Y )$, and the MI between $Z$ and the inputs $X$ during training: $I(X; Z)$.

\begin{figure}
    \centering
    \includegraphics[width=0.75\linewidth]{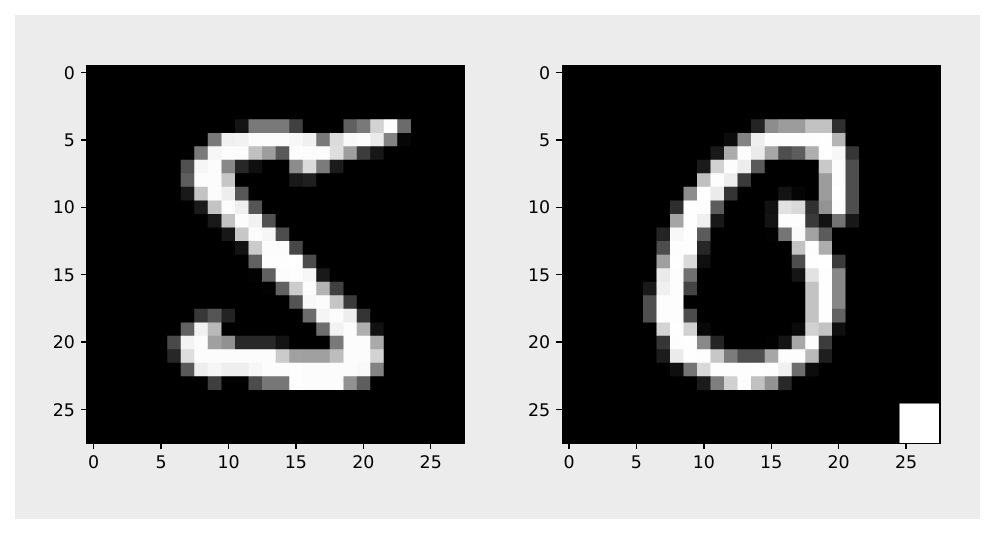}
    \caption{Sample images from MNIST dataset. A small patch is added to all images of even digits as a shortcut.}
    \label{fig:dataset}
\end{figure}

\section{Shortcuts and  information bottleneck}
Our hypothesis is that MI as measured during the evolution of a network's parameters can be used to monitor shortcut learning as it occurs. 
Shortcuts allow networks to learn a maximally compressed representation $Z$, i.e., $I(X; Z)$ is considerably 
reduced and thus can be used as a metric to monitor exploitation of shortcuts.

To support our hypothesis, we design controlled datasets containing spurious correlations and measure $I(X; Z)$ during the training evolution using the NTK. Using insights from the concept of a previously introduced ``mutual
information plane''~\cite{shwartz2017opening}, we can observe shortcut learning as it happens and find the time where the network
stops exploring the preferred (generalizable) solution space and latches onto the spurious signal.

\section{Experiments and observations}
\paragraph{Experimental Setup:} For each experiment, we plot $I(X; Z)$ and $I(Z; Y)$ w.r.t. time, generalization error/loss on clean data (without spurious correlations), and the information plane ($I(Z; Y)$ vs. $I(X; Z)$). Note that we calculate the upper bound of $I(X; Z)$, and are only interested in the training dynamics rather than the precise value of MI.

\paragraph{MNIST with synthetic shortcut:} We train a model to classify MNIST images into odd and even digits. We add a small white patch on one corner of \emph{all even} digits of the MNIST training dataset as a spurious correlation (\autoref{fig:dataset}). The network can use the patch to accurately classify the images into odd and even.
We compare the MI during the training evolution on datasets with and without shortcuts.

\begin{figure}[t]
    \begin{subfigure}[t]{0.48\linewidth}
      \includegraphics[width=\textwidth]{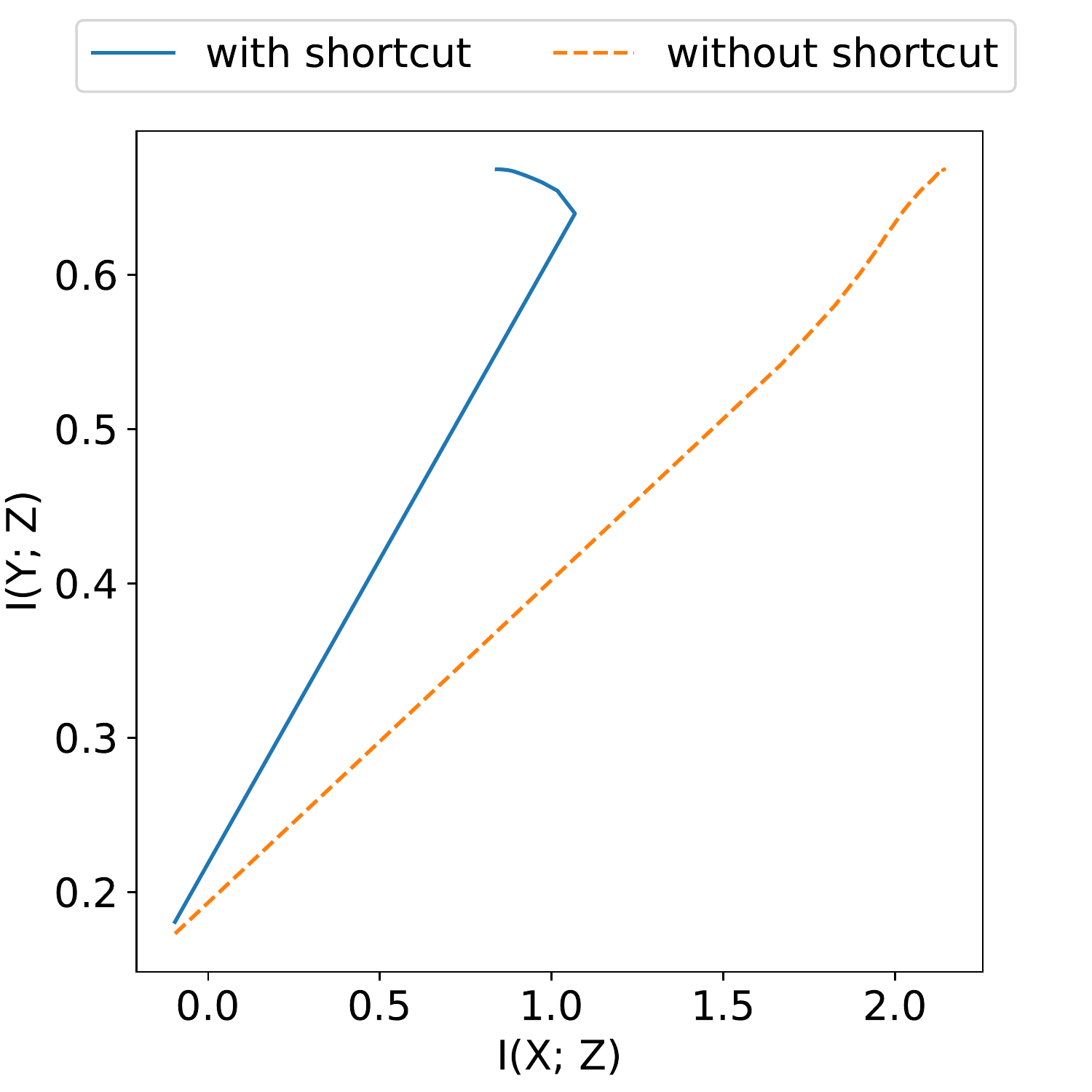}
      \caption{Information Plane}
      \label{fig:mnist_a}
    \end{subfigure}
    \hfill
    \begin{subfigure}[t]{0.48\linewidth}
      \includegraphics[width=\textwidth]{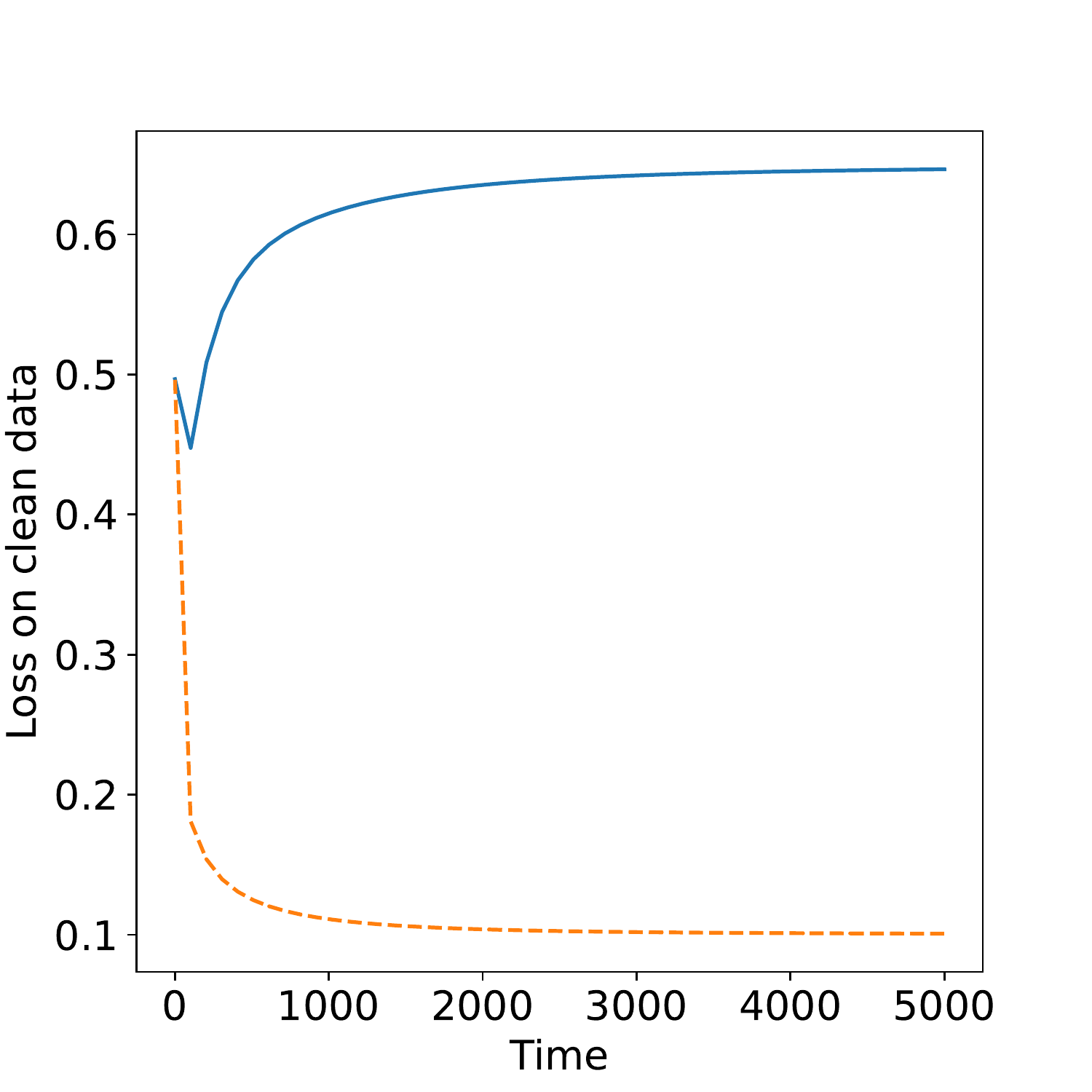}
      \caption{Clean Test Loss}
       \label{fig:mnist_b}
    \end{subfigure}
    
    \begin{subfigure}[t]{0.48\linewidth}
      \includegraphics[width=\textwidth]{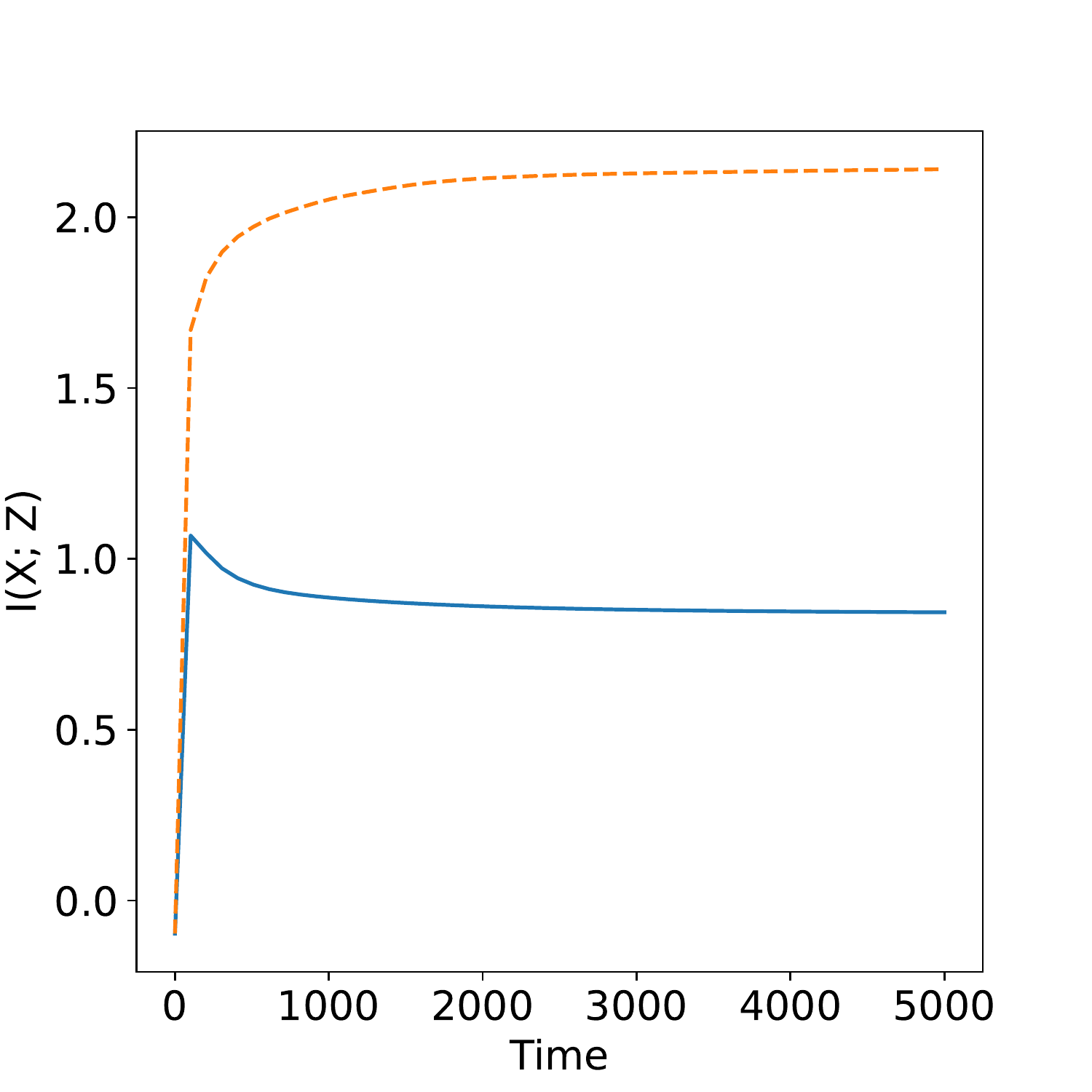}
      \caption{\(I(X; Z)\)}
      \label{fig:mnist_c}
    \end{subfigure}
    \hfill
    \begin{subfigure}[t]{0.48\linewidth}
      \includegraphics[width=\textwidth]{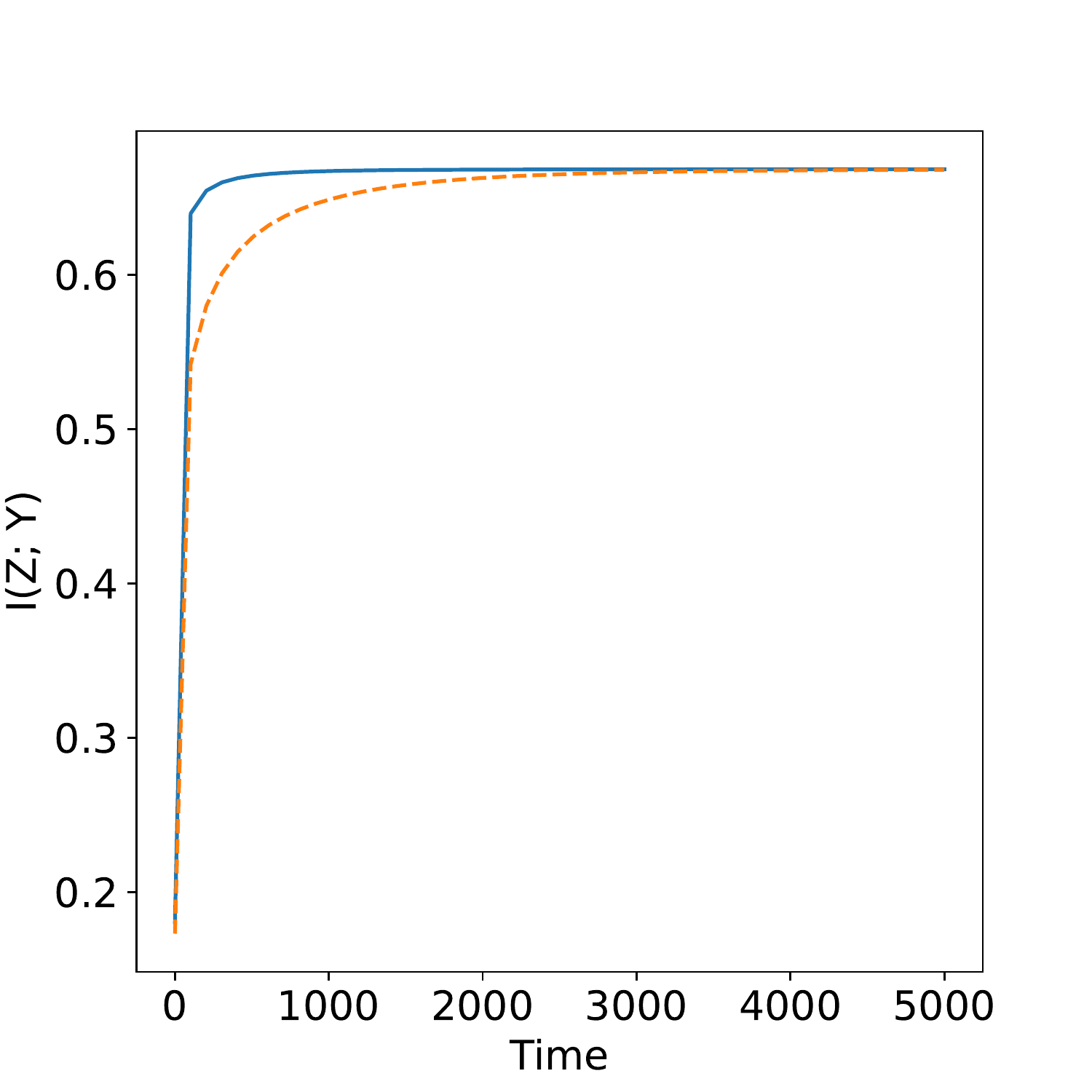}
      \caption{\(I(Z; Y)\)}
       \label{fig:mnist_d}
    \end{subfigure}
    
    \caption{Comparison of training evolution with and without synthetic shortcuts on MNIST dataset. An upper bound on MI is plotted.
    (a) Information plane plotting $I(X; Z)$ versus $I(Z; Y)$ during training. (b) Loss w.r.t.~the clean test data without shortcuts---the solid blue line corresponds to a model trained with shortcuts, the broken orange line is without shortcuts. (c) Plot of $I(X; Z)$ versus training time step. (d) Plot of $I(Y; Z)$ during the training evolution. Animated GIF of the plot can be viewed~\href{https://drive.google.com/file/d/1T3mf2h2B9cnNt4G0dQJABazc6DO67Coz/view?usp=sharing}{here}. }
    \label{fig:mnist}
\end{figure}



In \autoref{fig:mnist_a} and \autoref{fig:mnist_c}, we observe that the mutual information $I(X; Z)$ increases initially during training but then latches onto the shortcuts, after which the mutual information decreases sharply. 
It can also be observed in \autoref{fig:mnist_b} that generalization error increases after the point at which MI starts to decrease, indicating that the network explores the more optimal region of the solution space in the initial training epochs before discovering shortcuts. This is consistent with the findings of Schwartz-Ziv~and~Tishby~\yrcite{shwartz2017opening} about the behaviour of SGD. In the initial phase, SGD explores the multidimensional space of solutions.
When it begins converging, it arrives at the diffusion phase in which the network learns to compress~\cite{shwartz2017opening}. In both settings, the model achieves high training accuracy, i.e., $I(Z; Y)$~(\autoref{fig:mnist_d}) but the difference in test set loss is striking (\autoref{fig:mnist_b}).

\begin{figure}
    \centering
    \includegraphics[width=0.3\textwidth]{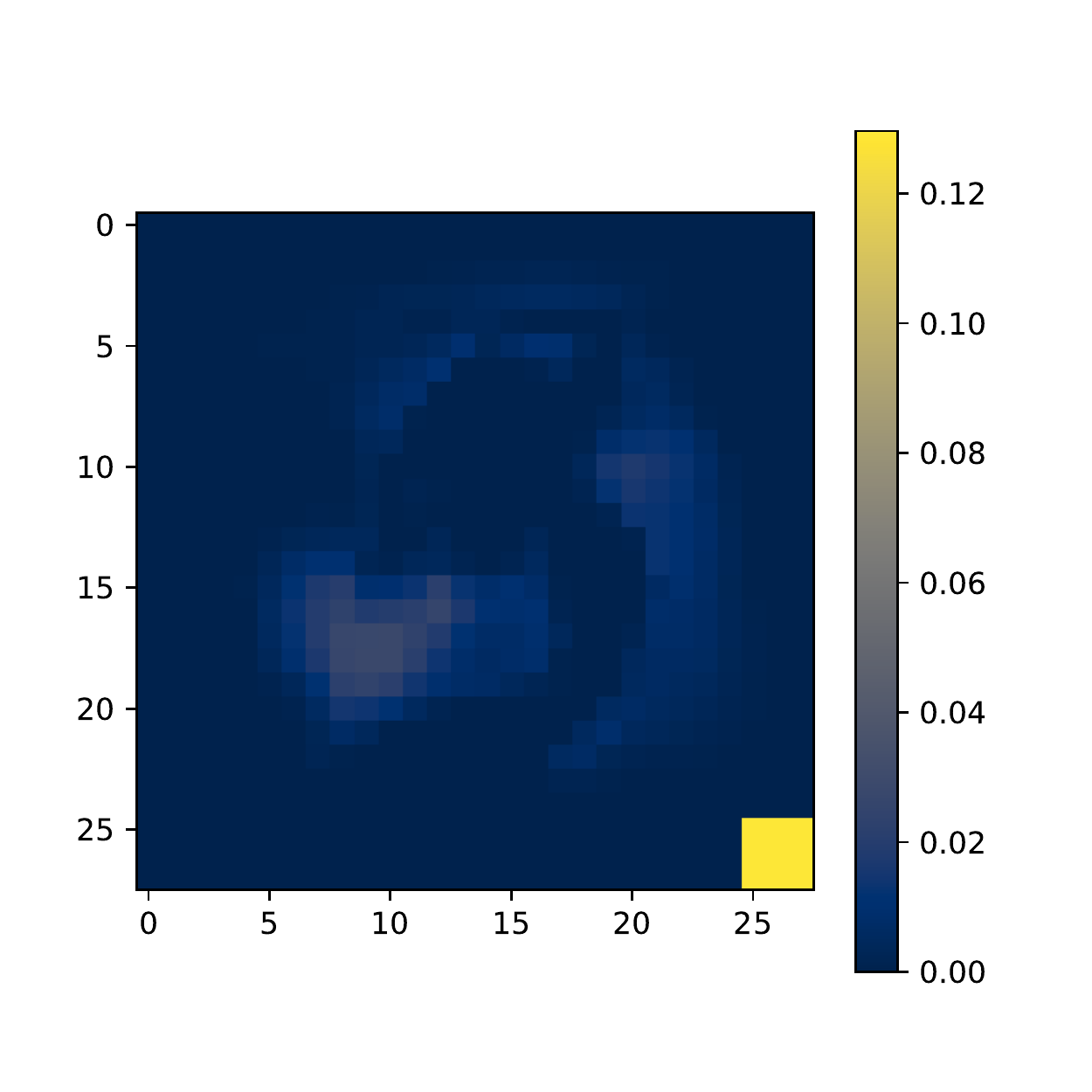}
    \vspace{-5mm}
    \caption{Saliency map of model trained on MNIST images with a small patch on even images as a shortcut.}
    \label{fig:visualization}
\end{figure}
\vspace{-2mm}

\paragraph{Visualisation:} To visually verify that the network is learning the shortcut in the above experiment, we generate a saliency map. 
We use the finite-difference estimation to find the gradient of the class probability w.r.t. to the input pixels. The network predominantly uses the shortcut in the image to predict the class (\autoref{fig:visualization}).

\begin{figure}[t]
    \begin{subfigure}[t]{0.48\linewidth}
      \includegraphics[width=\textwidth]{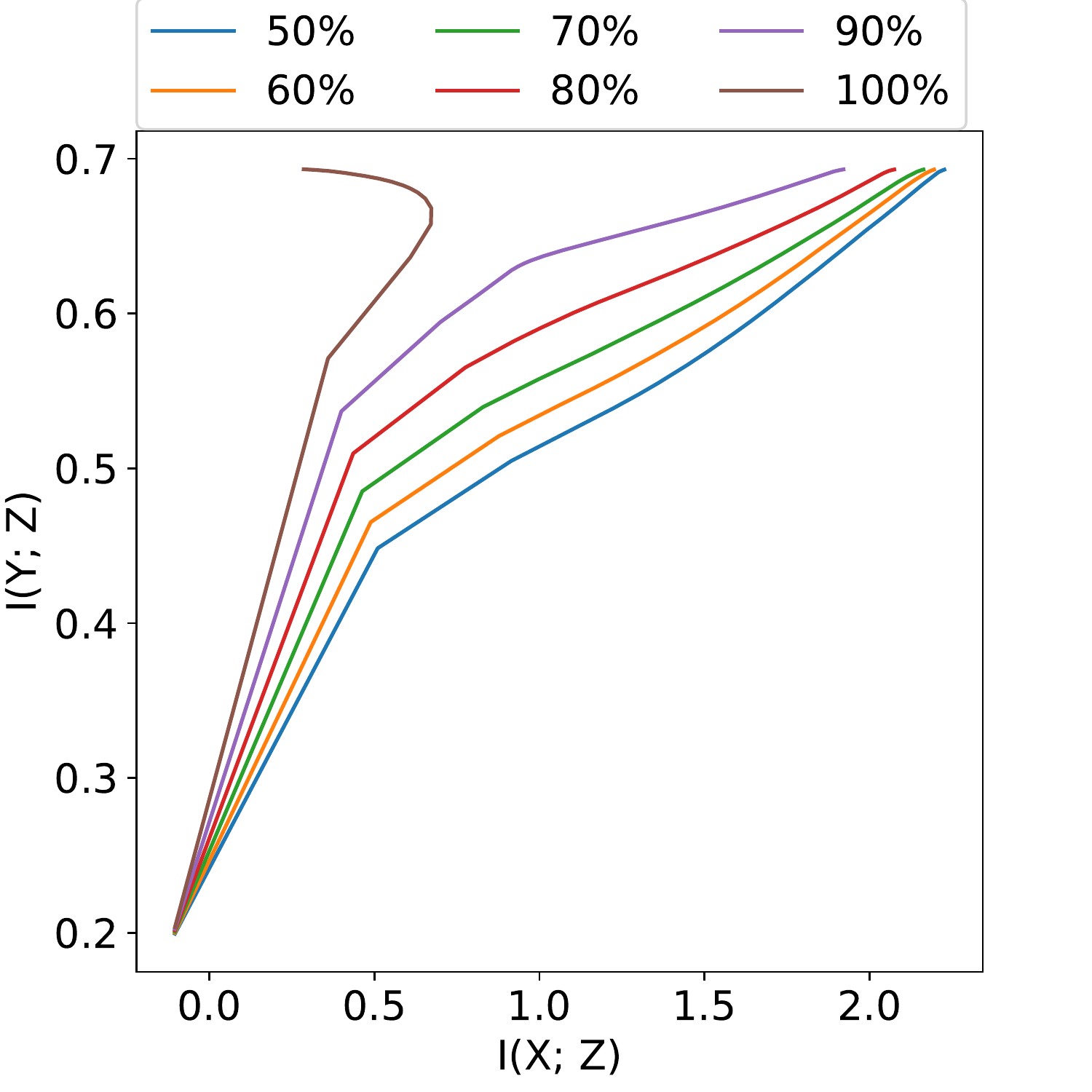}
      \caption{Information Plane}
      \label{fig:mnist_effectiveness_a}
    \end{subfigure}
    \hfill
    \begin{subfigure}[t]{0.48\linewidth}
      \includegraphics[width=\textwidth]{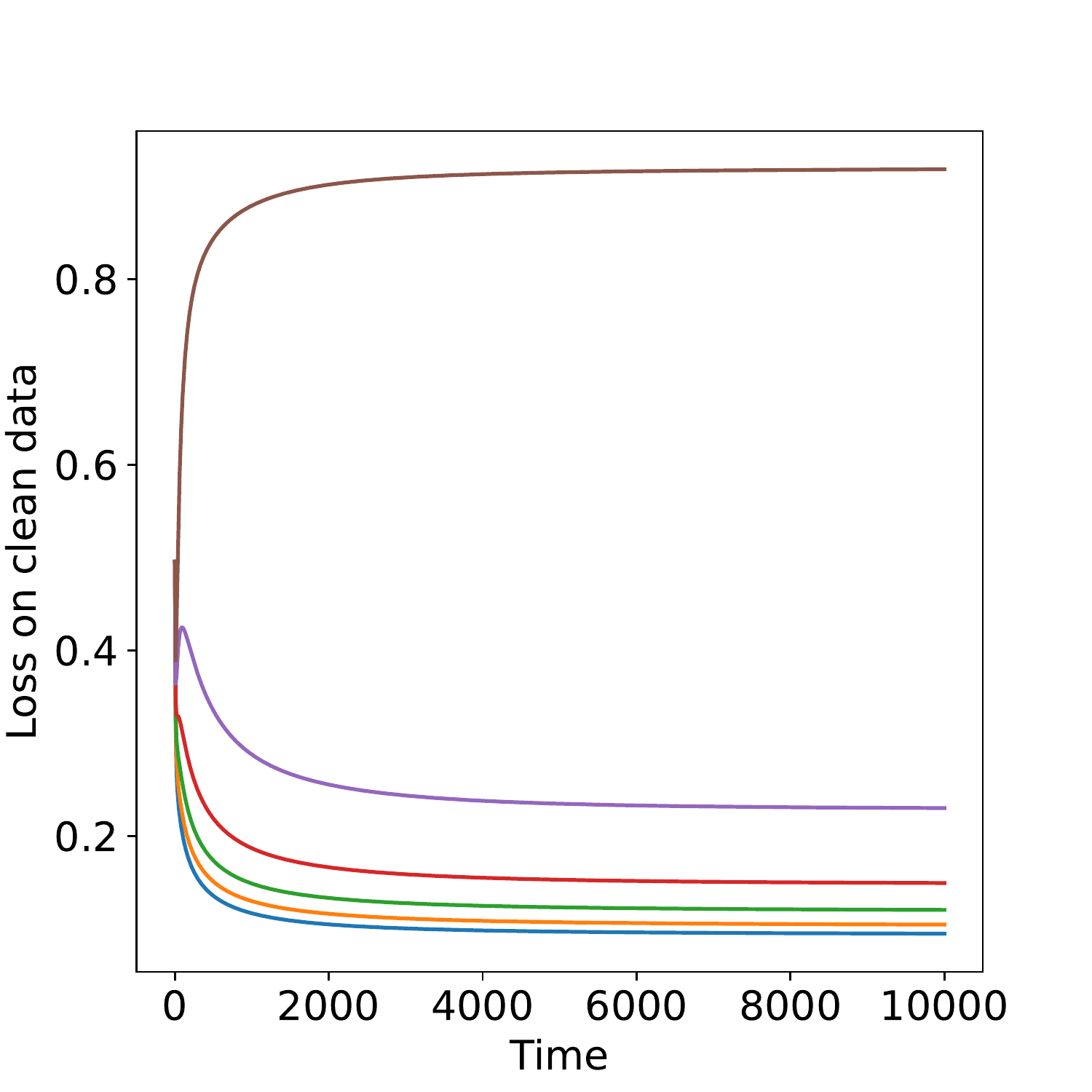}
      \caption{Clean Test Loss}
       \label{fig:mnist_effectiveness_b}
    \end{subfigure}
    
    \begin{subfigure}[t]{0.48\linewidth}
      \includegraphics[width=\textwidth]{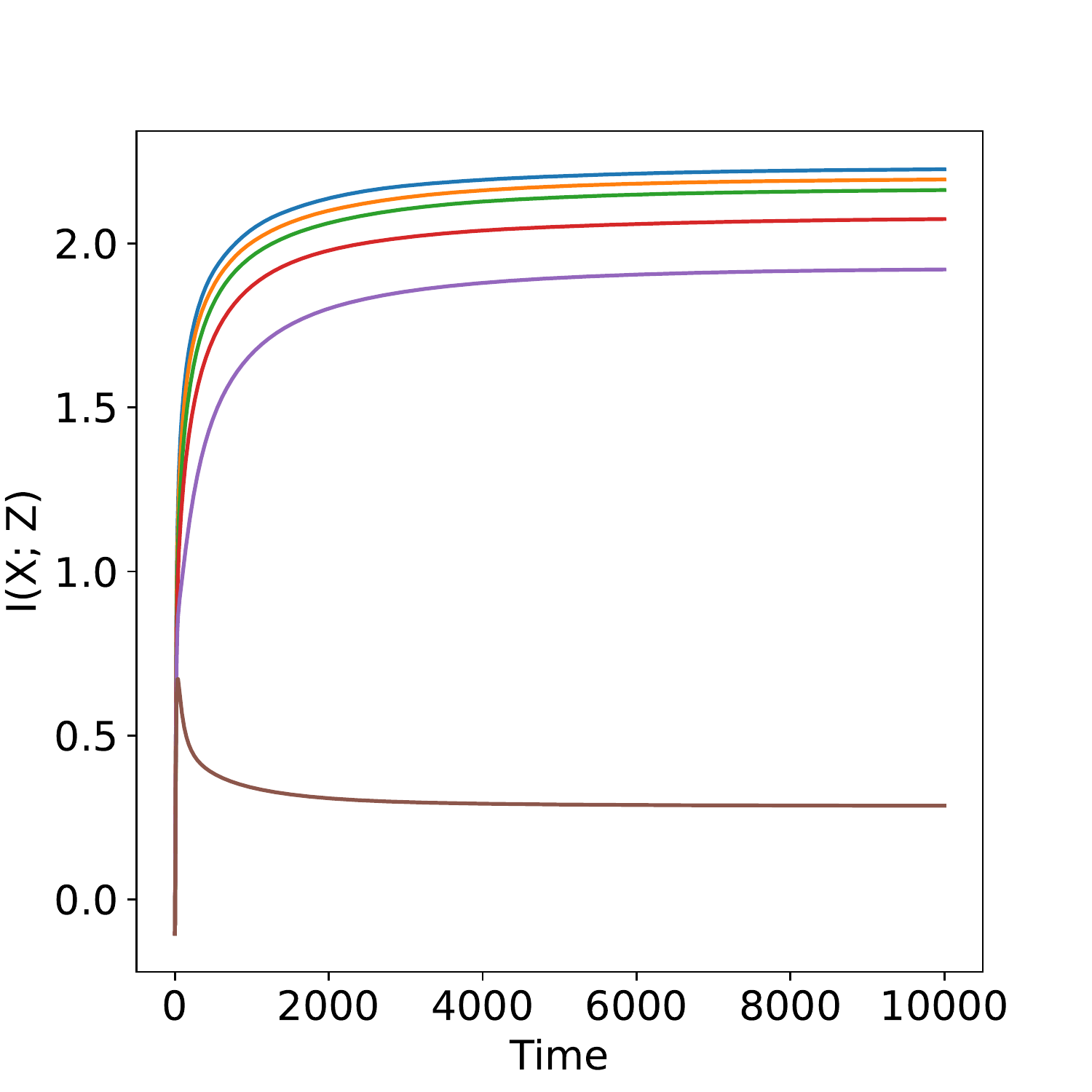}
      \caption{\(I(X; Z)\)}
      \label{fig:mnist_effectiveness_c}
    \end{subfigure}
    \hfill
    \begin{subfigure}[t]{0.48\linewidth}
      \includegraphics[width=\textwidth]{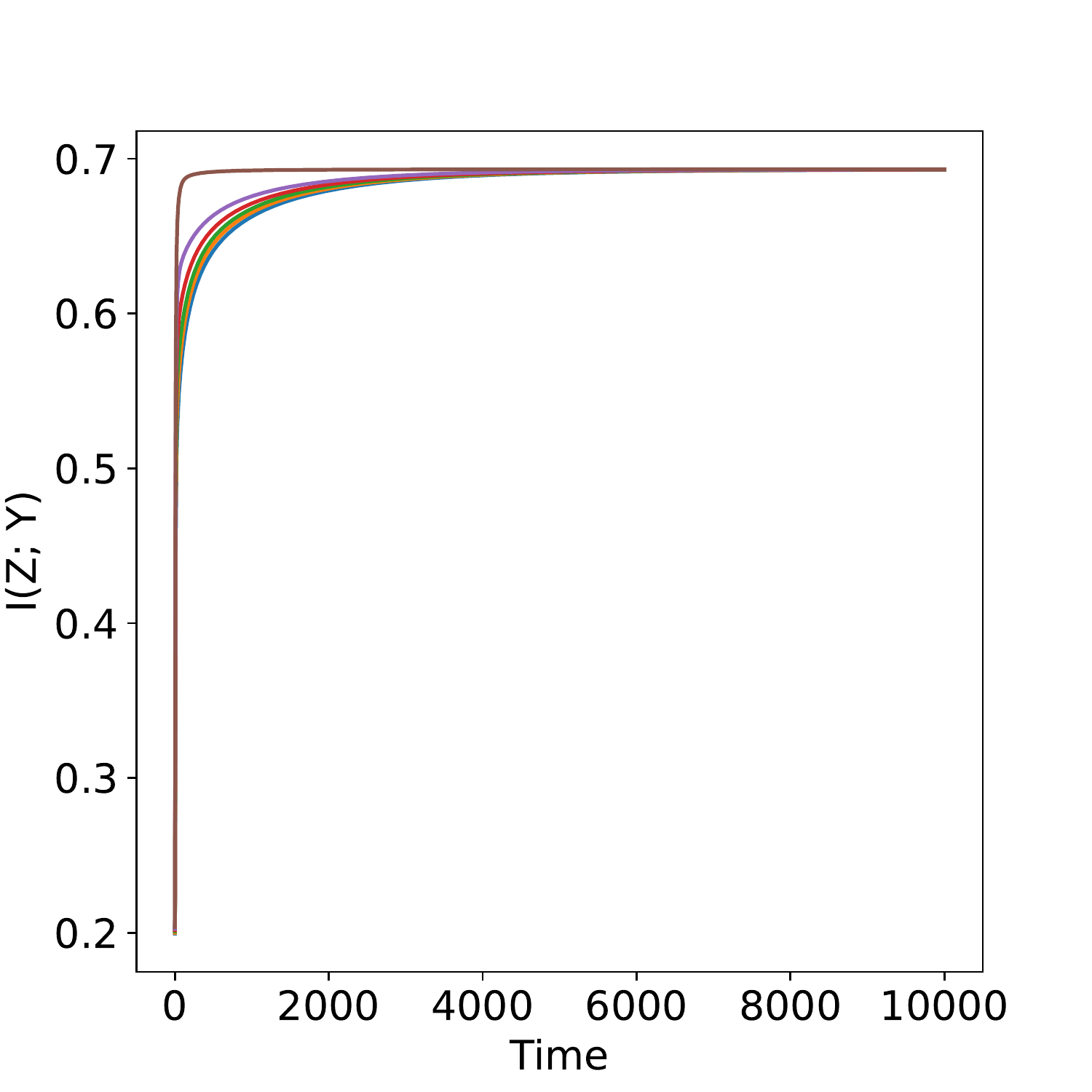}
      \caption{\(I(Z; Y)\)}
       \label{fig:mnist_effectiveness_d}
    \end{subfigure}
    
    \caption{Effect of shortcut effectiveness on the MI trajectory. Shortcut is added to different percentages of even images in each experiment. With increase in shortcut effectiveness, MI  converges to a lower value. Animated GIF of the plot can be viewed~\href{https://drive.google.com/file/d/1TTIdgISvLzFHdgq1Pj6VQVW27Ntx3diz/view?usp=sharing}{here}.}
    \label{fig:mnist_effectiveness}

\end{figure}

\paragraph{Effect of partially correlated shortcuts:} In real-world data, shortcuts are often partially correlated with the output, i.e., the model cannot classify with $100\%$ accuracy using only the shortcuts. 
To understand the effect of partially correlated shortcuts on the training dynamics, we construct different datasets with varying degrees of shortcut efficacy. Instead of corrupting all the even images, we add a small white patch on one corner only to a specific percentage (50\%--100\%) of even images. We plot the MI for varying degrees of corruption in \autoref{fig:mnist_effectiveness} for 1000 training points sampled uniformly.

We observe that as the effectiveness of the shortcut increases, $I(X; Z)$ converges to a lower value indicating the ability to perform more compression. We also note an interesting behaviour during the training evolution: when the shortcut is partially correlated, MI does not decrease significantly as compared to the 100\% effective shortcut.
We speculate that, while in these cases (e.g.~80--90\% shortcut efficacy) the model is able to recover some generalization ability, its ability to discover high generalization solutions is irrevocably deteriorated once it discovers the minima corresponding to the shortcut solution.

\begin{figure}[t]
    \begin{subfigure}[t]{0.48\linewidth}
      \includegraphics[width=\textwidth]{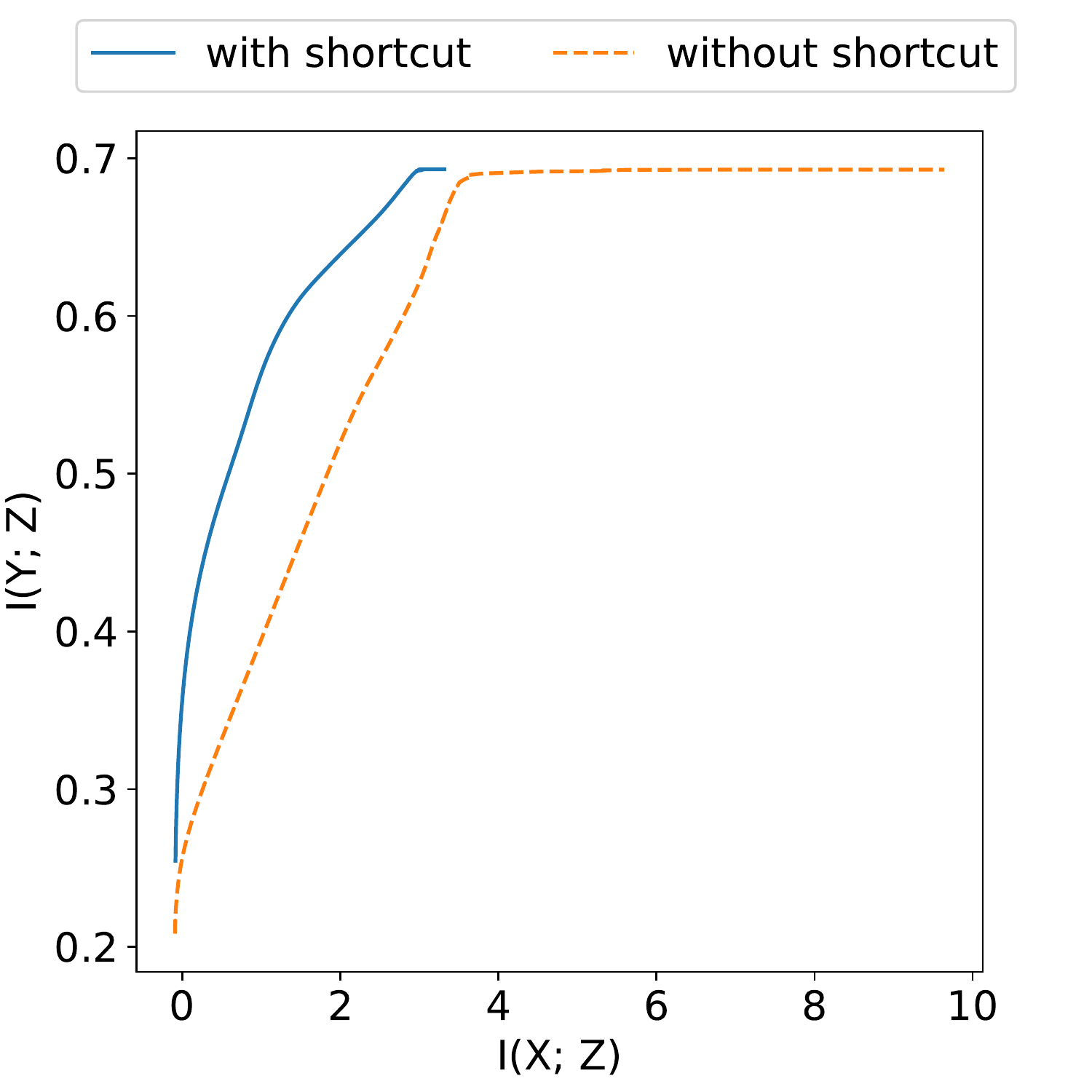}
      \caption{Information Plane}
      \label{fig:celeba_a}
    \end{subfigure}
    \hfill
    \begin{subfigure}[t]{0.48\linewidth}
      \includegraphics[width=\textwidth]{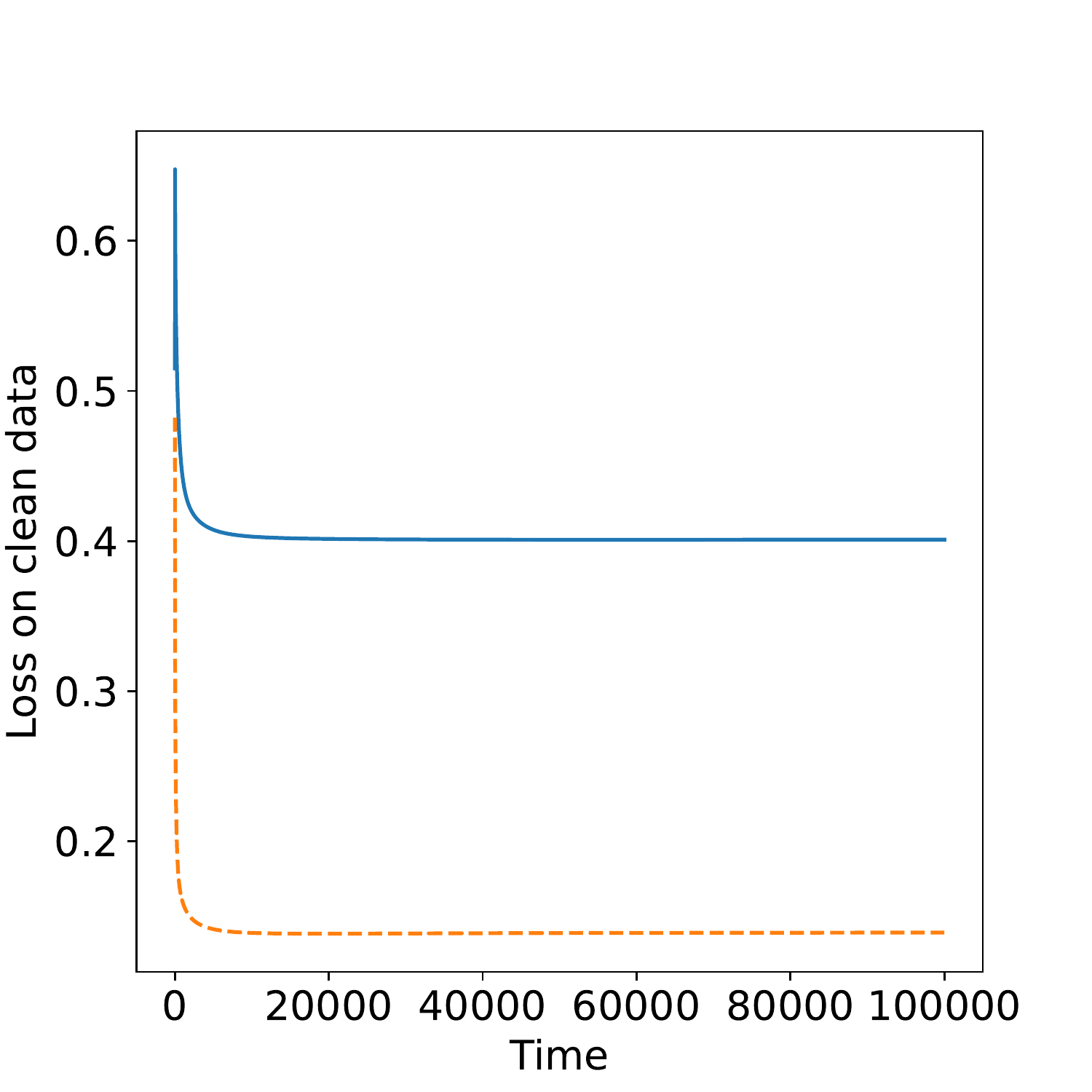}
      \caption{Clean Test Loss}
       \label{fig:celeba_b}
    \end{subfigure}
    
    \begin{subfigure}[t]{0.48\linewidth}
      \includegraphics[width=\textwidth]{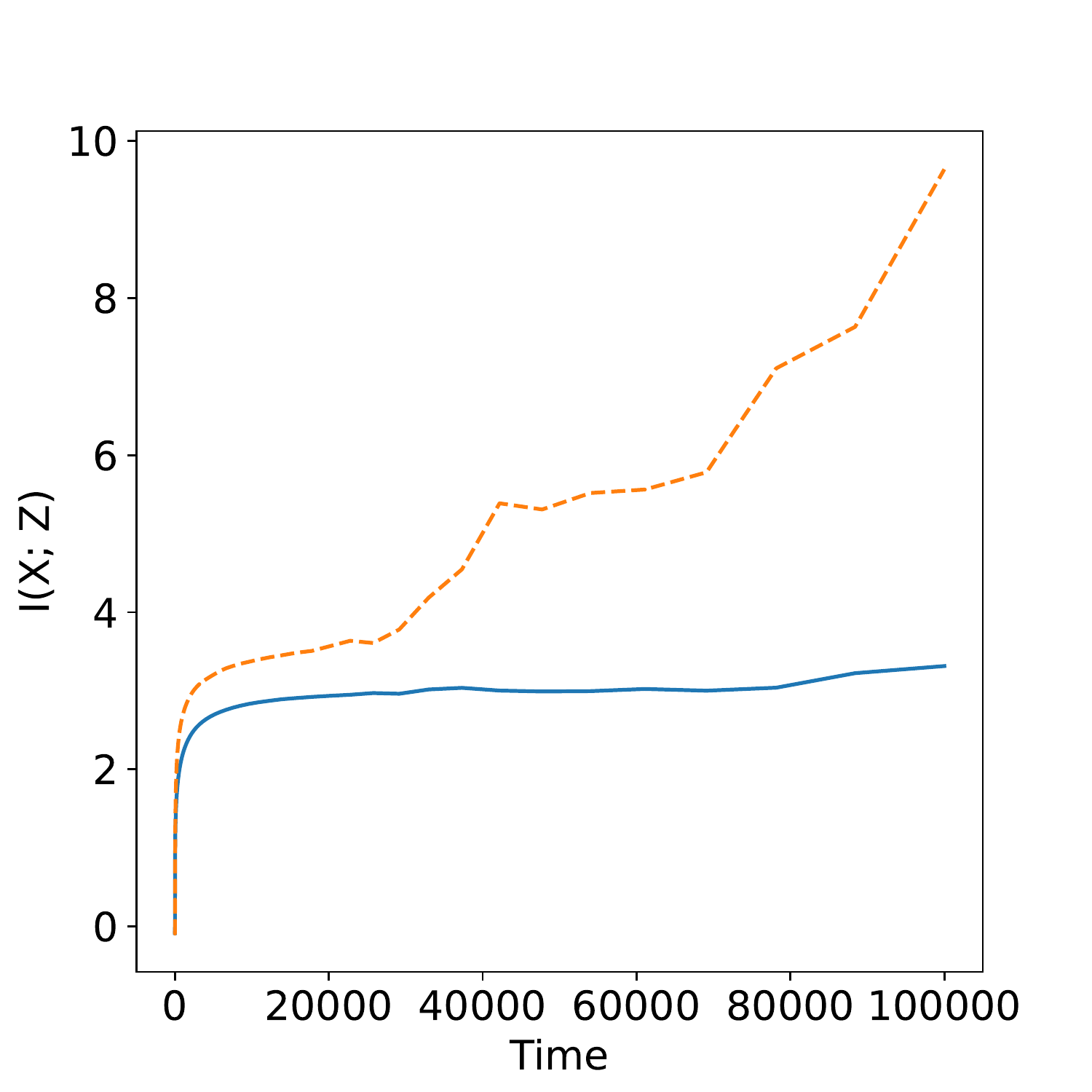}
      \caption{\(I(X; Z)\)}
      \label{fig:celeba_c}
    \end{subfigure}
    \hfill
    \begin{subfigure}[t]{0.48\linewidth}
      \includegraphics[width=\textwidth]{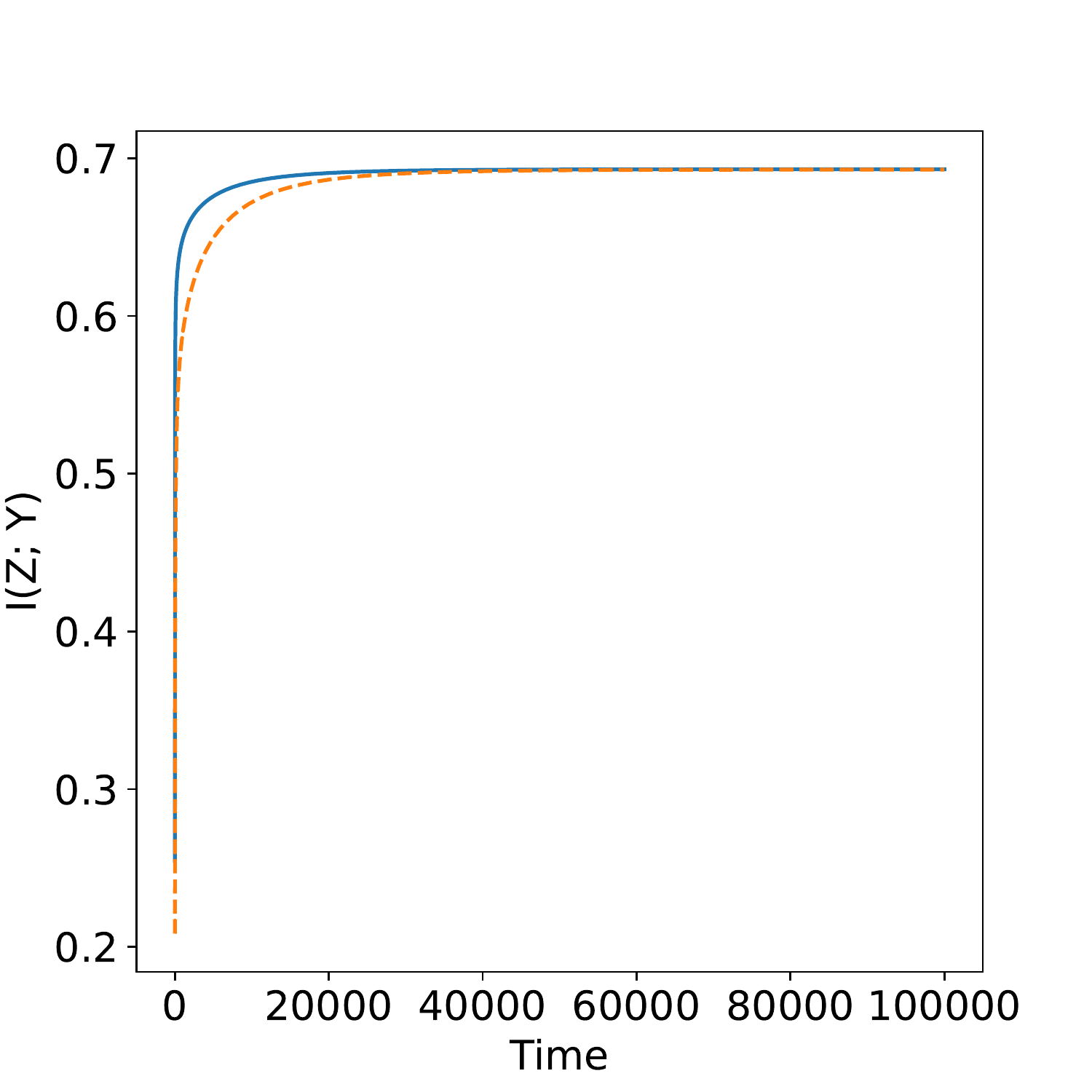}
      \caption{\(I(Z; Y)\)}
       \label{fig:celeba_d}
    \end{subfigure}
    
    \caption{Mutual information profile for CelebA dataset with hair attribute as shortcut. Plot of $I(X; Z)$ in (a) and (c) show that shortcuts results in reduced $I(X; Z)$. 
     Animated GIF of the plot can be viewed~\href{https://drive.google.com/file/d/1Y38iYfzvu7a4CP26kkCU-3OZ_wCjKTyv/view?usp=sharing}{here}.}
    \label{fig:celeba}

\end{figure}

\vspace{-2mm}
\paragraph{CelebA with natural shortcuts:} We also test our hypothesis on a dataset containing natural shortcuts. We curate images from the CelebA dataset such that all images tagged as male have the ``black hair color'' attribute, while images tagged as female have the ``blonde hair color'' attribute. We train the network to classify facial images into the male and female categories\footnote{The CelebA dataset only provides binary labels and we do not know how the gender attribute was assigned. Therefore it should be considered as nothing more than an arbitrary class in this experiment.}, while the network may use hair color as a shortcut for accurately classifying the images. Since the dataset is not controlled, some images from both classes have a black color in different parts of the image (background, clothes, etc.), reducing the effectiveness of the hair color attribute. We plot the MI trajectory with and without shortcuts for 100 sample training points sampled uniformly in $\log$ scale (\autoref{fig:celeba}).
We observe MI profile similar to  \autoref{fig:mnist_effectiveness}. On data without the shortcut, MI increases consistently, while in the presence of shortcuts, MI converges to a lower value, therefore validating our hypothesis on real-world data with natural shortcuts.

\begin{figure}[!b]
    \begin{subfigure}[t]{0.44\linewidth}
      \includegraphics[width=\textwidth]{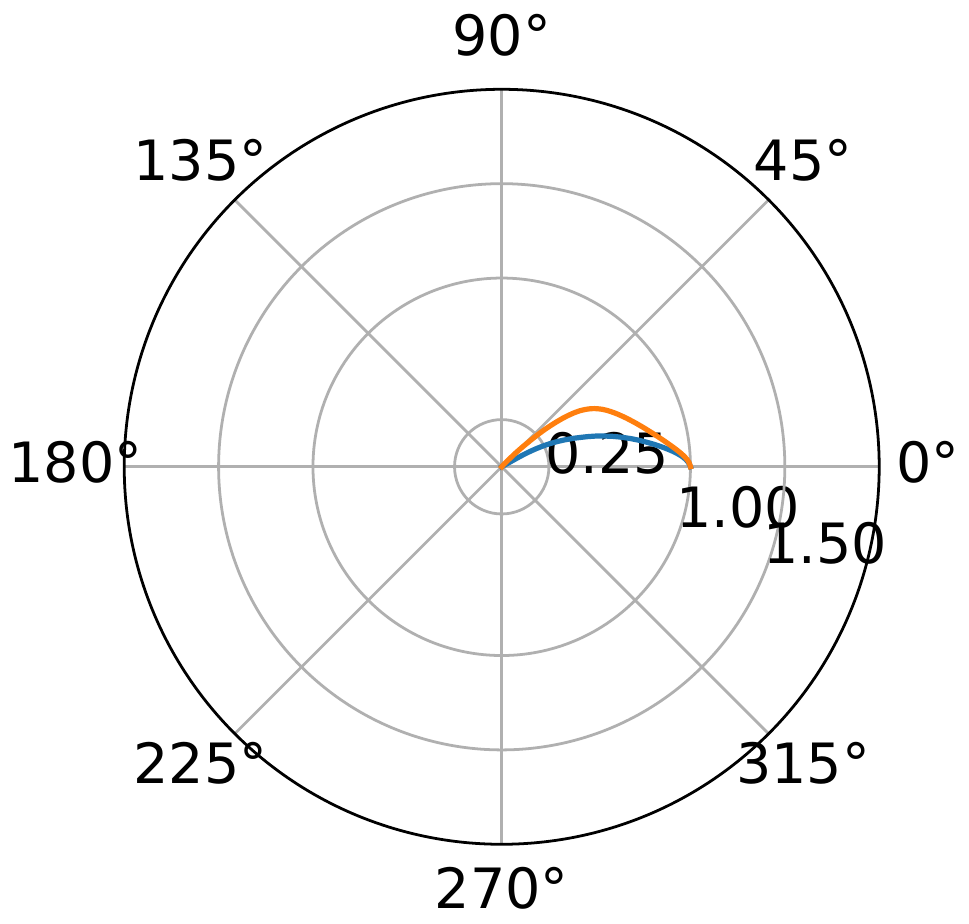}
      \caption{Radial plot}
    \end{subfigure}
   \hfill 
    \begin{subfigure}[t]{0.55\linewidth}
      \includegraphics[width=\textwidth]{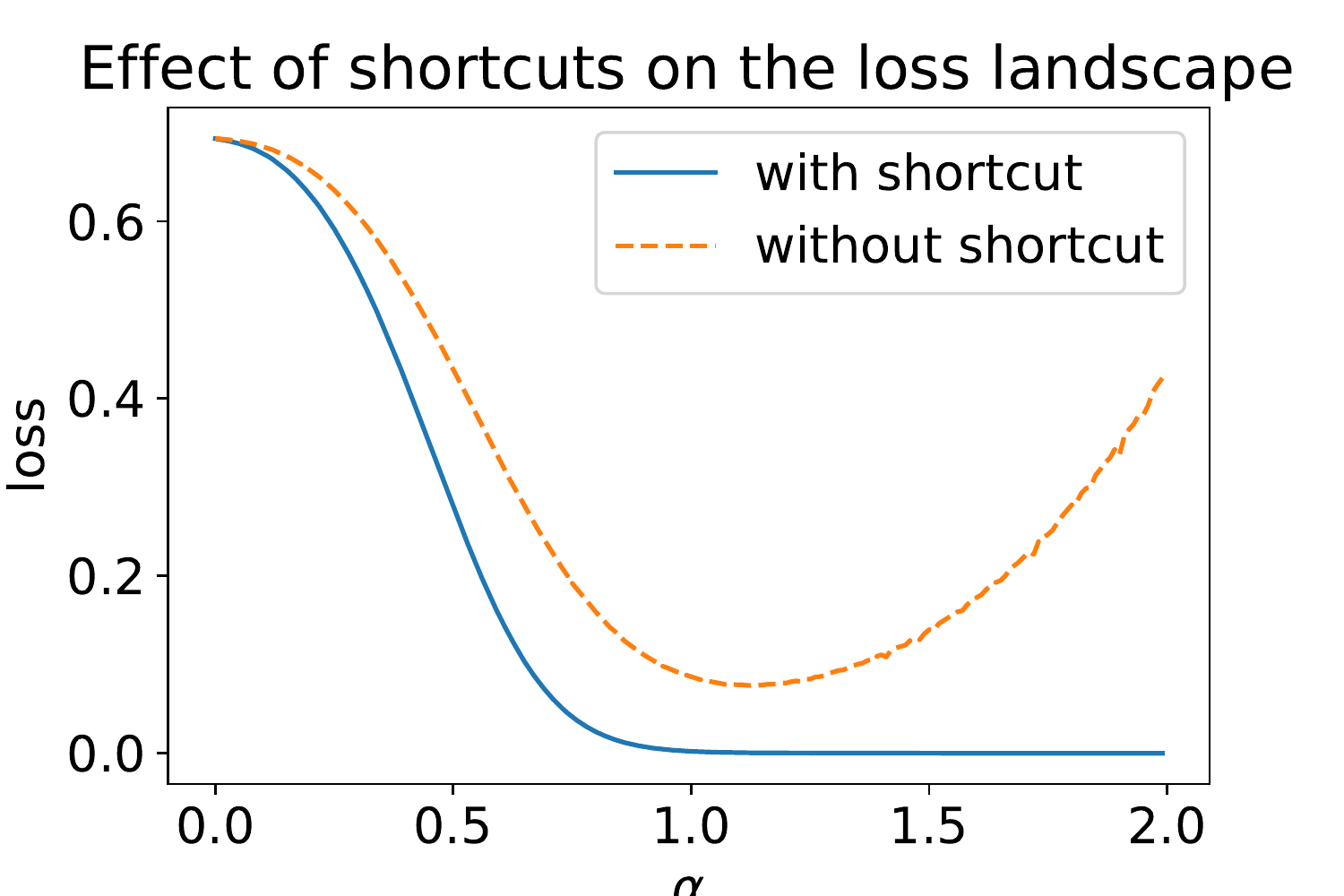}
      \caption{Loss landscape visualization}
    \end{subfigure}
    \caption{Visualization of loss landscape on MNIST dataset. (a). polar coordinates ($r_t, \phi_t)$ measuring the deviation from the linear path between initialisation and converged parameters in the weight space during the optimization. (b). 1-D visualization of loss landscape.
    }
    \label{fig:loss_landscape}
\end{figure}

\paragraph{Effect of shortcut on the loss landscape:}
We visualize the loss landscape of neural networks to understand the effect of shortcuts on the optimization trajectory. We plot loss along a linear path connecting the initial parameter $\theta_{\text{o}}$  and converged parameter $\theta^{*}$ in the weight space~\cite{goodfellow2014qualitatively} and polar coordinates ($r_t, \phi_t$) plot measuring the deviation from the linear line between $\theta_i$ and $\theta^{*}$ (\autoref{fig:loss_landscape}). We parameterize the line with $\alpha$ such that 
$\theta = (1-\alpha) \theta_i + \alpha  \theta^{*}$. 
Polar coordinates can be calculated using
$r_t = \frac{||\triangle \theta_t||}{||\triangle \theta_{\text{o}}||}$ and 
$\phi_t = \arccos{ \frac{\triangle \theta_t \times \triangle \theta_{\text{o}}} {||\triangle \theta_t|| \times ||\triangle \theta_{\text{o}}||}  }$,
where $\triangle \theta_t = \theta_t - \theta^{*}$.
We observe that the loss landscape around $\theta^{*}$ in the case of shortcuts is surprisingly flat as compared to the valley-like shape for a model trained on data not containing shortcuts using the MNIST dataset. The polar plot shows that the optimizer deviates less from the linear trajectory when trained with shortcuts.

\section{Conclusion and future work}
In this work, we sought to understand why networks tend to to learn shortcuts through 
the lens of the information bottleneck method. 
We showed that mutual information can be used to monitor training dynamics w.r.t.~shortcut learning without using any domain knowledge; this is an advantage compared to methods adapted by the interpretable ML literature, where domain knowledge is required. However, we used the NTK to estimate mutual information, limiting our approach to infinite-width neural networks. In the future, we will address this limitation and develop methods to avoid shortcut learning using tractable MI estimators that place less severe constraints on the model architecture~\cite{selby2022learning, gabrie2018entropy}.
We are also interested in exploring the relationship between the curvature of the solution minima and shortcut learning.

\bibliography{ref_agno, ntk, ref, angus_for_adnan}

\bibliographystyle{icml2022}



\end{document}